\documentclass[sigconf]{acmart}
\renewcommand\footnotetextcopyrightpermission[1]{}
\usepackage{graphicx}
\usepackage{subcaption}
\usepackage{float}
\usepackage{amsfonts}
\usepackage{bbm}  
\usepackage{multirow}
\usepackage{booktabs}
\usepackage{bm}
\def\pt{p_\textrm{t}}

\usepackage{bbm,nicefrac}
\usepackage{colortbl}

\AtBeginDocument{%
  }

\setcopyright{acmlicensed}
\copyrightyear{2025}
\acmYear{2025}
\acmDOI{XXXXXXX.XXXXXXX}
\acmConference[Conference acronym '25]{Make sure to enter the correct
  conference title from your rights confirmation email}{June 03--05,
  2025}{Woodstock, NY}

\acmSubmissionID{3875}

\begin{document}

\title{Referring Expression Instance Retrieval \\ and A Strong End-to-End Baseline}

\author{Xiangzhao Hao}
\email{haoxiangzhao2023@ia.ac.cn}
\affiliation{%
  \institution{Institute of Automation, Chinese Academy of Sciences}
  \city{Beijing}
  \country{China}
}

\author{Kuan Zhu}
\email{kuan.zhu@nlpr.ia.ac.cn}
\affiliation{%
  \institution{Institute of Automation, Chinese Academy of Sciences}
  \city{Beijing}
  \country{China}
}

\author{Hongyu Guo}
\email{23120800@bjtu.edu.cn}
\affiliation{%
  \institution{Institute of Automation, Chinese Academy of Sciences}
  \city{Beijing}
  \country{China}
}

\author{Haiyun Guo}
\email{haiyun.guo@nlpr.ia.ac.cn}
\affiliation{%
  \institution{Institute of Automation, Chinese Academy of Sciences}
  \city{Beijing}
  \country{China}
}

\author{Ning Jiang}
\email{ning.jiang02@msxf.com}
\affiliation{%
  \institution{Mashang Consumer Finance Co, Ltd}
  \city{Chongqing}
  \country{China}
}

\author{Quan Lu}
\email{quan.lu02@msxf.com}
\affiliation{%
  \institution{Mashang Consumer Finance Co, Ltd}
  \city{Chongqing}
  \country{China}
}

\author{Ming Tang}
\email{tangm@nlpr.ia.ac.cn}
\affiliation{%
  \institution{Institute of Automation, Chinese Academy of Sciences}
  \city{Beijing}
  \country{China}
}

\author{Jinqiao Wang}
\email{jqwang@nlpr.ia.ac.cn}
\affiliation{%
  \institution{Institute of Automation, Chinese Academy of Sciences}
  \city{Beijing}
  \country{China}
}


\renewcommand{\shortauthors}{Xiangzhao et al.}

\begin{abstract}
Using natural language to query visual information is a fundamental need in real-world applications, which has inspired a wide range of vision-language tasks. Text-Image Retrieval (TIR) retrieves a target image from a gallery based on an image-level description, while Referring Expression Comprehension (REC) localizes a target object within a given image using an instance-level description. However, real-world applications often present more complex demands. Users typically query an instance-level description across a large gallery and expect to receive both relevant image and the corresponding instance location. In such scenarios, TIR struggles with fine-grained descriptions and object-level localization, while REC is limited in its ability to efficiently search large galleries and lacks an effective ranking mechanism. In this paper, we introduce a new task called \textbf{Referring Expression Instance Retrieval (REIR)}, which supports both instance-level retrieval and localization based on fine-grained referring expressions. First, we propose a large-scale benchmark for REIR, named \textbf{REIRCOCO}, constructed by prompting advanced vision-language models to generate high-quality referring expressions for instances in the MSCOCO and RefCOCO datasets.
Second, we present a baseline method, \textbf{Contrastive Language-Instance Alignment with Relation Experts (CLARE)}, which employs a dual-stream architecture to address REIR in an end-to-end manner. Given a referring expression, the textual branch encodes it into a query embedding, enhanced by a Mix of Relation Experts (MORE) module designed to better capture inter-instance relationships. The visual branch detects candidate objects and extracts their instance-level visual features. The most similar candidate to the query is selected for bounding box prediction. CLARE is first trained on object detection and REC datasets to establish initial grounding capabilities, then optimized via Contrastive Language-Instance Alignment (CLIA) for improved retrieval across images.
Experimental results demonstrate that CLARE outperforms existing methods on the REIR benchmark and generalizes well to both TIR and REC tasks, showcasing its effectiveness and versatility. Our code and data will be released at \href{https://haoxiangzhao12138.github.io/REIR/}{this link}.
\end{abstract}

\maketitle

\begin{figure*}[t]
  \centering
    \includegraphics[width=0.95\textwidth]{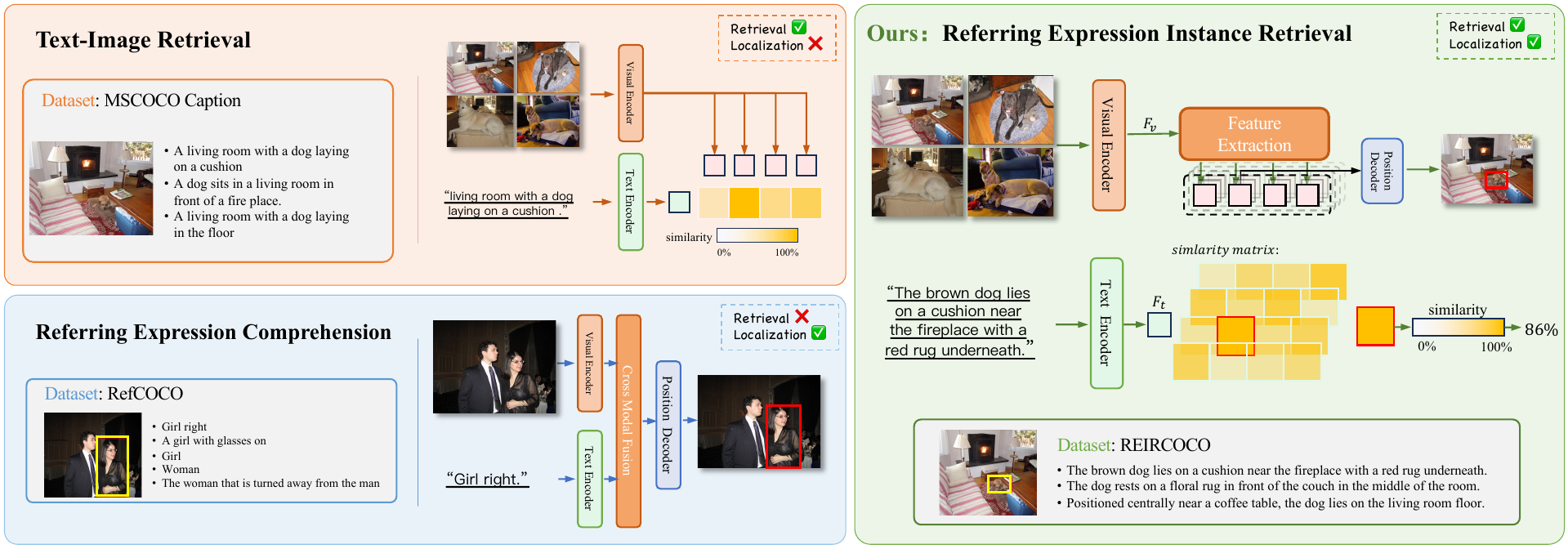}
        \caption{ Comparison of three vision-language tasks and their datasets. REIR enables end-to-end instance-level retrieval and localization, addressing the limitations of Text-Image retrieval and Referring Expression Comprehension.}
    \Description{}
    \label{fig:1}
\end{figure*}
\section{Introduction}
Querying visual information using natural language is a fundamental and widely demanded capability in vision-language systems.
Existing efforts in this area are often categorized into two representative tasks based on the granularity of the textual query and the scope of visual search.
\textbf{Text-Image Retrieval (TIR)} aims to retrieve relevant images from a gallery using image-level captions. Recent TIR methods typically adopt large-scale contrastive learning between paired images and texts~\cite{clip, siglip, evaclip, GLIP}, enabling strong global retrieval performance.
\textbf{Referring Expression Comprehension (REC)}, in contrast, focuses on localizing a specific object instance within a single image based on fine-grained referring expressions.
State-of-the-art REC models usually rely on early cross-modal fusion modules to align image and text features, followed by decoder heads to predict the location of the referred object~\cite{lavt, cris, simvg, yang2019dynamic}.
These differences in task scope and methodology are illustrated in Figure~\ref{fig:1}.

\textbf{However}, real-world applications such as surveillance retrieval, image forensics, and personal photo search often demand a hybrid capability: given an instance-level natural language expression, the system should directly retrieve and localize the corresponding object instance from a large image gallery.
Such expressions typically describe salient objects using both appearance and relational context, e.g., \textit{The brown dog lies on a cushion near the fireplace with a red rug underneath}.
Unfortunately, while existing TIR methods are efficient, they lack localization capabilities and fail to identify specific object instances. REC models, in contrast, offer precise localization within a single image but rely on expensive cross-modal feature fusion and are impractical for gallery-wide retrieval.
Moreover, neither task provides a unified evaluation protocol for jointly assessing retrieval and grounding performance.

To address these limitations, we propose a new task: \textbf{Referring Expression Instance Retrieval (REIR)}.  
REIR requires models to \textit{directly retrieve and localize the referred object instance from a gallery of images}, based on a fine-grained natural language expression.  
To support this task, we introduce a new dataset, \textbf{REIRCOCO}, which provides high-quality instance-level expressions paired with precise bounding box annotations.
As illustrated in Figure~\ref{fig:1}, existing datasets are not suitable for this task.  
TIR datasets like MSCOCO Captions~\cite{mscococaption} only contain image-level descriptions (e.g., \textit{A living room with a dog lying on a cushion}) without object-level annotations, while REC datasets such as RefCOCO~\cite{refcoco,refcoco2} rely on short, ambiguous expressions (e.g., \textit{Girl}) that can refer to multiple targets in gallery settings. 
In contrast, REIRCOCO offers fine-grained and unambiguous expressions (e.g., \textit{The brown dog lies on a cushion near the fireplace with a red rug underneath}), uniquely grounding specific object instances even in cluttered scenes.
We construct REIRCOCO using an automated pipeline with two stages: generation and filtering.  
In the generation stage, we prompt a vision-language model to produce contextualized referring expressions for each object in the MSCOCO detection dataset~\cite{mscoco} and RefCOCO dataset.  
Each prompt incorporates the object’s attributes, spatial relations, and surrounding context to encourage relationally rich and instance-specific descriptions.  
In the filtering stage, a separate language model evaluates expression quality, removing samples that are ambiguous, inaccurate, or non-discriminative.  
The resulting dataset contains over 30,000 images and 200,000 uniquely annotated object instances, each paired with several fine-grained referring expressions that unambiguously correspond to a single target. 

To address the REIR task, we propose a new end-to-end model named \textbf{CLARE} (Contrastive Language-Instance Alignment with Relation Experts). CLARE adopts a dual-stream architecture that decouples text and image encoding for efficient cross-modal alignment. Given a referring expression, the text encoder produces a sentence-level embedding, which is refined by a \textbf{Mix of Relation Experts (MORE)} module to enhance relational semantics. In parallel, the visual encoder extracts instance-level object features, which encapsulate not only the appearance attributes of each object but also their spatial and contextual relationships within the image. To align the two modalities, we introduce \textbf{Contrastive Language-Instance Alignment (CLIA)} module, a contrastive learning objective that matches textual queries with object candidates across images using the SigLIP loss~\cite{siglip}. 
This enables CLARE to effectively align instance-level visual features with referring expressions, allowing the model to identify the most relevant object among all candidate instances across the gallery, thereby achieving simultaneous retrieval and localization.

To evaluate our approach, we construct strong baselines by combining and adapting representative TIR and REC methods in the REIR setting, and these adapted methods perform poorly. We compare these baselines with our proposed CLARE under standardized settings and observe that CLARE achieves consistently superior performance on the REIR task. Extensive ablation studies further validate the effectiveness of both our proposed model and the REIRCOCO dataset. Furthermore, CLARE also performs competitively on traditional REC and TIR benchmarks, demonstrating strong generalization across vision-language tasks.
\begin{figure*}[t]
  \centering
  \includegraphics[width=0.85\linewidth]{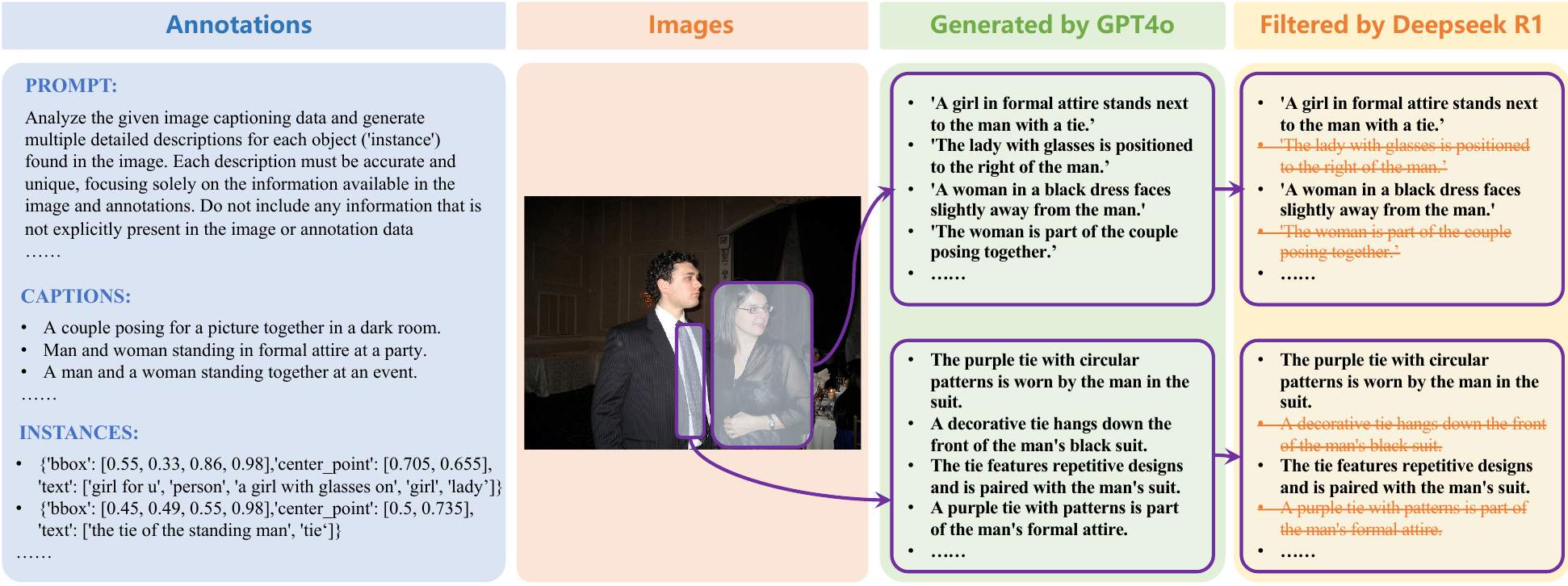}
  \caption{Overview of the REIRCOCO dataset construction pipeline. GPT-4o produces context-rich and uniquely grounded referring expressions for annotated instances using structured prompts. DeepSeek R1 verifies expression quality and removes ambiguous or low-informative samples.}
  \Description{}
  \label{fig:2}
\end{figure*}
Our main contributions are summarized as follows:
\begin{itemize}
\item We introduce a new task, \textbf{Referring Expression Instance Retrieval (REIR)}, which better reflects real-world vision-language demands and extends the scope of traditional retrieval and grounding tasks.

\item We construct a large-scale dataset, \textbf{REIRCOCO}, where each object instance is annotated with a semantically rich instance-level referring expression. To the best of our knowledge, this is the first dataset specifically designed for instance-level retrieval and localization in open-world scenarios.

\item We propose \textbf{CLARE} (Contrastive Language-Instance Alignment with Relation Experts), an end-to-end framework that enables simultaneous retrieval and localization through instance-level language-object contrastive alignment.

\item We conduct extensive experiments on REIR, TIR, and REC benchmarks. CLARE achieves state-of-the-art performance on the REIR task, while also demonstrating strong generalization capabilities on traditional REC and TIR settings.
\end{itemize}

\section{Related Works}
\subsection{Referring Expression Comprehension}

Referring Expression Comprehension (REC) aims to localize a specific object within a given image based on a natural language expression.
Existing approaches are typically classified into two-stage and one-stage paradigms, mainly distinguished by whether they perform early fusion of visual and textual features. \textbf{Two-stage methods}~\cite{mattnet, cmatt, dga, rvg, nmtree, MobileNetv3} detect object proposals using pre-trained object detectors such as Faster R-CNN~\cite{fasterrcnn}, and then compute cross-modal similarity between each region and the referring expression. These methods benefit from modularity and interpretability, but generally lack deep feature fusion and end-to-end optimization. As a result, they struggle to model complex relationships or semantic nuances, especially in open-world or gallery-level settings. \textbf{One-stage methods}\cite{lavt,cris, seqtr, transvg, mdetr} bypass the proposal stage by jointly encoding images and text through early fusion in a unified architecture. With the rise of vision-language transformers~\cite{transformer} and large-scale pretrained multimodal models (MLLMs)~\cite{shikra, ferret, Qwen, deepseekvl2}, one-stage methods have demonstrated superior performance across multiple benchmarks, largely due to their stronger representation learning and global reasoning capacity.

However, most REC methods rely on early feature fusion, which requires recomputing cross-modal features for each text-image pair. This makes them inefficient for large-scale gallery retrieval. In contrast, our method adopts a contrastive language-instance alignment strategy that enables effective grounding \textit{without requiring explicit feature fusion}. As a result, it achieves competitive or even superior grounding performance while maintaining scalability across large image galleries.

\subsection{Text-Image Retrieval}
Text-Image Retrieval (TIR) aims to retrieve the most relevant image from a gallery based on a natural language query. Early approaches~\cite{Faghri_Fleet_Kiros_Fidler_2017, Frome_Corrado_Shlens_Bengio_Dean_Ranzato_Mikolov_2013, Nam_Ha_Kim_2017} constructed joint embedding spaces using handcrafted or CNN-based features, but these methods lacked semantic understanding and showed poor generalization. To address these limitations, some later works~\cite{Lee_Chen_Hua_Hu_He_2018, Liu_Mao_Liu_Zhang_Wang_Zhang_2019, Wang_Liu_Li_Sheng_Yan_Wang_Shao_2019, Wang_Yang_Qian_Ma_Lu_Li_Fan_2019} introduced object detectors to extract region-level features, enabling word-region matching. For example, SCAN~\cite{Lee_Chen_Hua_Hu_He_2018} aligns detected image objects with textual fragments to improve retrieval precision. The current dominant paradigm is Vision-Language Pretraining (VLP), which leverages large-scale image-text pairs and self-supervised objectives~\cite{Li_Zhang_Li_Li_Fu_2019, Kipf_Welling_2016, Ballakur_Arya_2020}. These models are typically divided into two categories: one-stream and two-stream architectures. One-stream models~\cite{Kholy_Ahmed_Gan_Cheng_Liu_2020, Gan_Chen_Fu_Zhu_Cheng_Liu_2020, Kim_Son_Kim_2021, Zhang_Mao_Wang_Zhang} jointly encode image regions and text tokens through a unified transformer for dense cross-modal interaction. In contrast, two-stream models~\cite{GLIP, BLIP, BEITv3, Huang_Zeng_Huang_Liu_Fu_Fu_2021, Radford_Kim_Hallacy_Ramesh_Goh_Agarwal_Sastry_Amanda_Mishkin_Clark_et1} such as CLIP~\cite{clip}, EVA-CLIP~\cite{evaclip}, and SigLIP~\cite{siglip}, encode images and texts independently and align them through contrastive learning, enabling efficient large-scale retrieval.

Despite their effectiveness, most TIR models focus on image-level alignment and lack explicit modeling of object-level semantics. This limitation motivates us to extend contrastive alignment to the instance level, enabling retrieval and localization of fine-grained targets as required by REIR.
\begin{figure*}[t]
  \centering
  \includegraphics[width=0.8\linewidth]{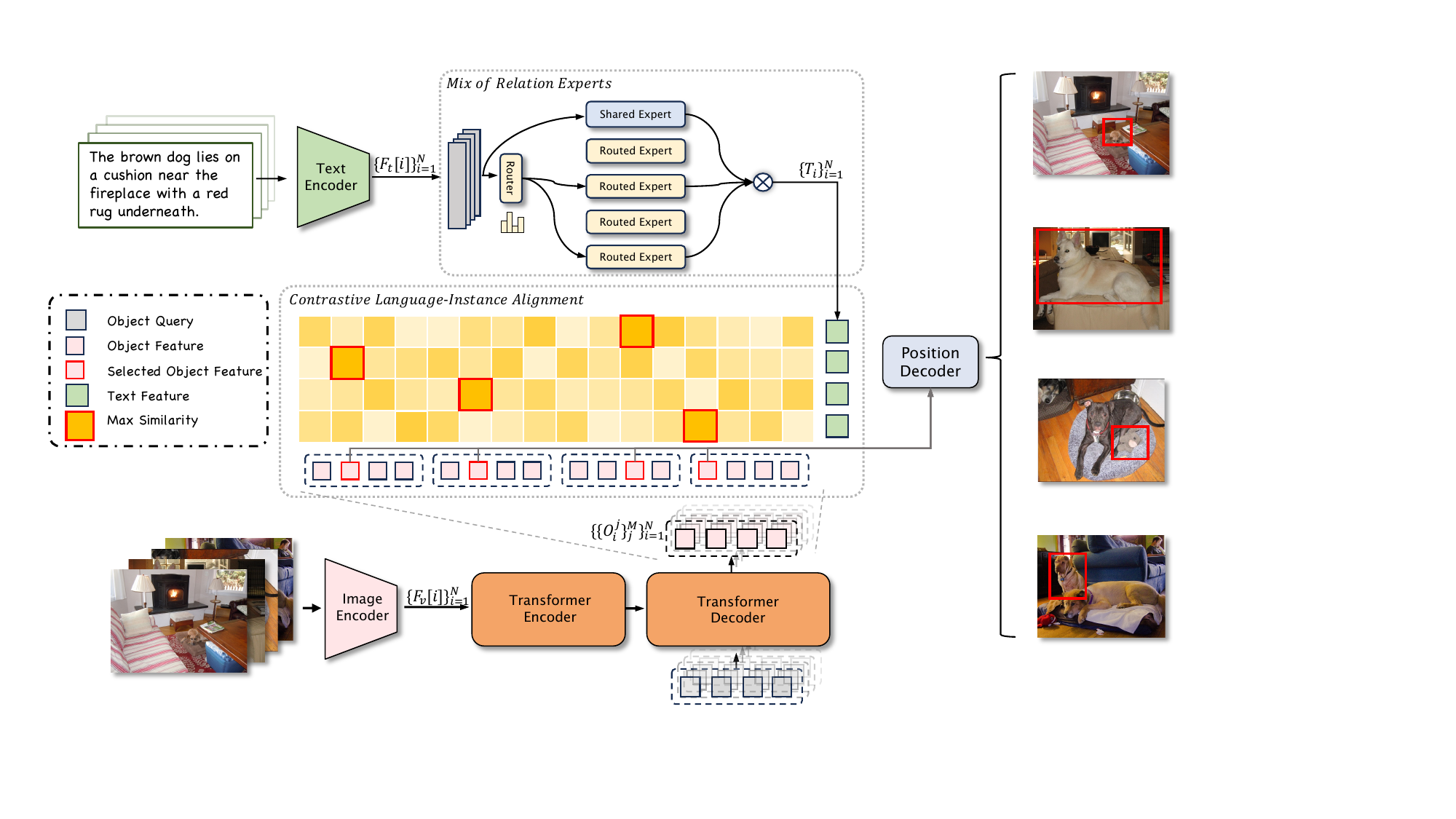}
  \caption{An overview of CLARE. The model encodes texts and images in parallel, then aligns referring expressions with instance-level object features via a Mix of Relation Experts (MORE) module and a Contrastive Language-Instance Alignment (CLIA) objective. This enables unified retrieval and localization across large-scale image galleries.}
  \Description{}
  \label{fig:3}
\end{figure*}
\section{Referring Expression Instance Retrieval}
\subsection{Task Definition}
\textbf{Referring Expression Instance Retrieval (REIR)} is a unified vision-language task that requires models to \textit{directly retrieve and localize} a referred object instance from a large-scale image gallery, given a natural language query.

Formally, given a referring expression $x_{\text{text}}$ and a gallery of $N$ images $\{x_{\text{img}}^i\}_{i=1}^N$, the model is required to predict a bounding box $B = [x, y, w, h]$ for the referred object, which need to lie within the correct image $x_{\text{img}}^i$ that contains the target instance. The model receives no prior information about which image contains the object, it needs to reason over all candidates to jointly determine both \textit{where} and \textit{what} to localize.

For instance, given the query “The brown dog lies on a cushion near the fireplace with a red rug underneath,” the model must retrieve the correct image from a large gallery and localize the exact dog being described. This is particularly challenging because many distractors may exist across the gallery—such as dogs lying on sofas, dogs outdoors, or images containing multiple dogs. The model must accurately match fine-grained appearance and contextual cues to locate the referred instance. As illustrated in Figure~\ref{fig:1}, this requires both precise retrieval and localization in the presence of semantically similar but incorrect candidates.


\subsection{A New Dataset: REIRCOCO}

To support the proposed REIR task, we construct a large-scale dataset named \textbf{REIRCOCO}, specifically designed for instance-level retrieval and localization. Unlike existing datasets, REIRCOCO features fine-grained and uniquely aligned textual descriptions for individual object instances, ensuring that each expression corresponds to exactly one target. This eliminates ambiguous references and negative truths, enabling accurate instance-level evaluation.

As illustrated in Figure~\ref{fig:2}, REIRCOCO is built through a scalable two-stage pipeline powered by vision-language foundation models: \textit{generation} and \textit{filtering}. In the \textbf{generation stage}, we begin with object annotations from the COCO detection and RefCOCO datasets. For each target instance, we construct structured prompts that include its bounding box, category label, the image’s global caption, and contextual information about surrounding objects. These prompts are then passed to GPT-4o~\cite{gpt4o} to generate 5–10 diverse referring expressions per instance. To reduce the risk of \textit{negative truths}—i.e., cases where a query unintentionally matches multiple similar objects across different images—the prompts are designed to maximize referential uniqueness. In particular, the model is encouraged to incorporate relational cues and spatial context involving nearby entities (e.g., “the chair to the left of the woman in red”) so that the generated expressions uniquely describe the target instance within the broader visual scene. This design ensures that each query expression has only one valid match in the entire image gallery, aligning perfectly with the REIR task requirement.
In the \textbf{filtering stage}, we use DeepSeek R1~\cite{deepseekr1} to verify the quality of the generated descriptions. Each image is paired with a verification prompt that includes the caption and candidate expressions. The model is instructed to reject expressions that are ambiguous or semantically inconsistent with the image context, and to retain those that are specific, grounded, and discriminative. This automated filtering ensures that the final dataset contains only high-quality, retrieval-friendly expressions. See the Appendix for more details.

The resulting REIRCOCO dataset contains \textbf{30,106 images} and \textbf{215,835 object instances}, each annotated with multiple verified referring expressions—totaling \textbf{613,548 fine-grained descriptions}. Each expression is unambiguous and uniquely tied to a single instance, making the dataset well-suited for both retrieval and localization. Compared to prior datasets, REIRCOCO provides richer semantics, greater diversity, and stronger alignment with the REIR task objectives.

\section{Contrastive Language-Instance Alignment with Relation Experts}
\subsection{Architecture Overview}
As shown in Figure~\ref{fig:3}, our proposed model \textbf{CLARE} (\textit{Contrastive Language-Instance Alignment with Relation Experts}) is an end-to-end dual-stream design tailored for instance-level retrieval and localization. It decouples vision and language processing while maintaining strong cross-modal alignment capabilities.

Given a batch of $N$ image-text pairs $\{(x_{\text{text}}^i, x_{\text{img}}^i)\}_{i=1}^{N}$, the model processes each modality in parallel. On the visual side, each image $x_{\text{img}}^i$ is passed through SigLIP vision encoder~\cite{siglip} to produce a dense feature map $F_v^i \in \mathbb{R}^{H \times W \times C}$. A Deformable-DETR-based object extractor is then applied to generate $M$ object proposals and extract instance-level visual features $\{O_i^j\}_{j=1}^{M} \in \mathbb{R}^D$by combining local appearance with global context using multi-scale deformable attention and Transformer decoding. These features encode both the visual appearance of each candidate object and its contextual relations within the scene, enabling the model to distinguish between visually similar instances in different spatial configurations. Simultaneously, the referring expression $x_{\text{text}}^i$ is encoded into a sentence-level embedding $F_t^i$ using the text encoder. This embedding is refined by the \textbf{MORE} module (\textit{Mix of Relation Experts}) into $T_i \in \mathbb{R}^D$, which incorporates semantic, spatial, and inter-object relational cues. To perform instance-level alignment, we compute dot-product similarity scores between $T_i$ and all object features ${O_k^l}$ from the gallery. These scores are supervised by the proposed \textbf{CLIA} module (\textit{Contrastive Language-Instance Alignment}, which extends SigLIP’s sigmoid-based contrastive objective to align referring expressions and object features across images. This allows the model to learn a unified similarity space for retrieval and localization. In addition, a Focal Loss is applied to supervise intra-image candidate selection and encourage discriminative matching.

\subsection{Mix of Relation Experts}

Referring expressions often combine both appearance cues and complex inter-object relationships. For example, “the man” captures a category-level visual concept, while “the man beside the woman in black” introduces spatial and relational context essential for disambiguating similar objects. To model this diversity, we propose the \textbf{Mix of Relation Experts (MORE)} module, which dynamically routes the global textual feature $F_t^i$ through a mixture of experts.

Inspired by DeepSeek-MoE, the MORE module comprises two types of experts: \textit{routed} and \textbf{shared experts}. The shared experts are always active and encode general semantics common across all expressions, while the routed experts are selectively activated based on each expression to specialize in fine-grained reasoning, such as spatial configurations, inter-object interactions, and comparative relations.
Let $N_s$ and $N_r$ denote the number of shared and routed experts. The final text embedding is:
\begin{equation}
T_i = \sum_{j=1}^{N_s} \text{FFN}^{(s)}_j(F_t^i) + \sum_{k=1}^{N_r} \hat{g}_{k,i} \cdot \text{FFN}^{(r)}_k(F_t^i)
\end{equation}
Routing weights $\hat{g}_{k,i}$ are obtained via a GatingNet:
\begin{align}
    \mathbf{g}_i &= \text{GatingNet}(F_t^i), \\
    \hat{g}_{k,i} &= \begin{cases} g_{k,i}, & \text{if } g_{k,i} \in \text{Top-}K_r(\mathbf{g}_i) \\
    0, & \text{otherwise} \end{cases}
\end{align}
Only the top-$K_r$ routed experts are activated per query to promote specialization and reduce redundancy. This design allows the model to dynamically adapt to diverse expression types and better encode the compositional semantics required for fine-grained retrieval and localization.

\subsection{Contrastive Language-Instance Alignment}
\label{4.4}
To support gallery-wide instance retrieval, we extend the sigmoid-based contrastive objective from SigLIP~\cite{siglip} to enable alignment between referring expressions and object-level visual features. This module, \textbf{Contrastive Language-Instance Alignment (CLIA)}, casts cross-modal matching as a binary classification task over all expression-object pairs across images and batches.
Given a batch of $B$ referring expressions ${T_i}$ and all extracted object features ${O_k^l}$, we define the label $z_{i,k,l} = 1$ if the expression $T_i$ refers to object $O_k^l$, and $-1$ otherwise. The CLIA loss is defined as:
\begin{equation}
\resizebox{0.92\linewidth}{!}{$
\mathcal{L}_{\text{CLIA}} = - \frac{1}{B \cdot N} \sum_{i=1}^{B} \sum_{k=1}^{B} \sum_{l=1}^{N_k} \log \left( \frac{1}{1 + \exp  \left(z_{i,k,l} (- t  \langle T_i, O_k^l \rangle + b) \right)} \right)$}
\end{equation}
In which, $\langle \cdot , \cdot \rangle$ denotes dot-product similarity, $t$ is a learnable temperature, and $b$ is a bias scalar. 
By aligning each expression not only with objects in its paired image but also across other gallery images, CLIA encourages the model to learn globally discriminative instance-level features that reflect semantic, spatial, and relational consistency which enable robust retrieval and localization across large-scale datasets.

\subsection{Training and Inference}

\textbf{Training.} CLARE is trained in two phases. First, a pretraining phase builds basic recognition and alignment skills. Using COCO and RefCOCO, we apply Focal Loss and box regression to train the model without routed experts or CLIA.
\begin{equation}
\mathcal{L}_{\text{pretrain}} = \mathcal{L}_{\text{focal}} + \mathcal{L}_{\text{bbox}}
\end{equation}

Then, a fine-tuning stage activates the full MORE module and applies CLIA loss on REIRCOCO:
\begin{equation}
\mathcal{L}_{\text{finetune}} = \mathcal{L}_{\text{CLIA}} + \mathcal{L}_{\text{focal}} + \mathcal{L}_{\text{bbox}}
\end{equation}

\textbf{Inference.} CLARE supports scalable retrieval: instance-level features for all gallery images are precomputed. At test time, a query is encoded once and compared via dot product to all candidates, enabling efficient and accurate retrieval-localization without repeated fusion.
This modular structure combines the flexibility of two-stream design with the fine-grained reasoning power of contrastive and expert-based modeling, making CLARE highly effective for the REIR task. See the Appendix for more details.

\begin{table*}[t]
\centering
\caption{Performance Comparison of Different Methods on the REIR Benchmark}
\label{tab:1}
\resizebox{0.85\textwidth}{!}{
\small
\begin{tabular}{ll|ccc|ccc|ccc}
\toprule
\multirow{2}{*}{Methods} & 
& \multicolumn{3}{c|}{$\tau=0.5$} 
& \multicolumn{3}{c|}{$\tau=0.7$} 
& \multicolumn{3}{c}{$\tau=0.9$} \\
\cmidrule(){3-5} \cmidrule(){6-8} \cmidrule(){9-11}
& & BR@1 & BR@5 & BR@10 & BR@1 & BR@5 & BR@10 & BR@1 & BR@5 & BR@10 \\
\hline
\multicolumn{1}{l}{TIR Methods} & 
\multicolumn{1}{l}{REC Methods} & 
\multicolumn{9}{r}{\textbf{Two-Stage Method}} \\\hline
CLIP-ViT-B & deepseek-VL2-Tiny & 11.15 & 21.84 & 26.89 & 8.98 & 18.00 & 23.50 & 5.32 & 10.51 & 13.48 \\
CLIP-ViT-L & deepseek-VL2      & 13.31 & 26.45 & 32.79 & 12.85 & 25.41 & 31.53 & 11.02 & 21.42 & 26.50 \\
CLIP-ViT-L & simVG-ViT-L        &12.01 & 25.68 & 31.78 & 10.16 & 23.79 & 30.04 & 7.31 & 20.75 & 21.21 \\
EVA-CLIP-ViT-B & deepseek-VL2-Tiny & 12.82 & 24.58 & 29.52 & 11.56 & 21.08 & 25.18 & 6.05 & 12.42 & 14.80 \\
EVA-CLIP-ViT-L & deepseek-VL2      & 16.38 & 29.89 & 36.21 & 15.77 & 28.73 & 34.78 & 13.32 & 24.18 & 29.15 \\
EVA-CLIP-ViT-L & simVG-ViT-L       & 15.92 & 29.04 & 36.22 & 14.78 & 27.63 & 33.39 & 8.06 & 20.38 & 28.41 \\
SIGLIP-ViT-B & deepseek-VL2-Tiny   & 15.26 & 27.43 & 32.93 & 13.27 & 23.60 & 28.34 & 7.63 & 13.53 & 16.09 \\
SIGLIP-ViT-L & deepseek-VL2        & 18.52 & 33.22 & 39.48 & 17.82 & 31.90 & 37.84 & 15.04 & 26.74 & 31.50 \\
SIGLIP-ViT-L & simVG-ViT-L         & 17.77 & 33.22 & 38.71 & 15.10 & 31.42 & 35.99 & 8.74 & 25.27 & 27.41 \\
\hline
\multicolumn{11}{r}{\textbf{End to End Method}} \\
\hline
\multicolumn{2}{c|}{CLARE-ViT-B (ours)} & \underline{28.47} & \underline{49.50} & \underline{56.85} & \underline{25.96} & \underline{42.85} & \underline{48.57} & \underline{19.38} & \underline{31.53} & \underline{35.48} \\
\multicolumn{2}{c|}{CLARE-ViT-L (ours)} & \textbf{29.53} & \textbf{50.56} & \textbf{56.98} & \textbf{28.03} & \textbf{47.07} & \textbf{53.34} & \textbf{21.12} & \textbf{35.80} & \textbf{40.35} \\
\bottomrule
\end{tabular}
}
\end{table*}
\section{Experiment}
\subsection{Evaluation Metrics}
Evaluating Referring Expression Instance Retrieval (REIR) requires measuring both retrieval and localization performance within a unified framework. To achieve this, we adopt standard metrics from related tasks and introduce a novel metric tailored for REIR.

\textbf{REC Metric: Precision@0.5.}  
Referring Expression Comprehension (REC) methods typically use Precision@0.5 to evaluate localization accuracy. A prediction is considered correct if the Intersection over Union (IoU) between the predicted and ground-truth bounding boxes exceeds 0.5:
\begin{equation}
\text{IoU}(B_{\text{pred}}, B_{\text{gt}}) > 0.5
\end{equation}

\textbf{TIR Metric: Recall@k.}  
Text-to-Image Retrieval (TIR) methods are evaluated using Recall@k, which measures the fraction of queries for which the correct image appears in the top-$k$ retrieved results. Formally, it is defined as:
\begin{equation}
\text{Recall@}k = \frac{1}{N} \sum_{i=1}^{N} \mathbbm{1}[ \text{GT}(i) \in \text{Top-}k(i) ]
\end{equation}

\textbf{REIR Metric: BoxRecall@k (IoU>$\tau$).}
To jointly evaluate retrieval and localization in the REIR task, we adopt a unified metric called \textit{BoxRecall@k (IoU>$\tau$)}. A prediction is considered correct if the referred object instance is ranked among the top-$k$ candidates and its predicted bounding box sufficiently overlaps with the ground-truth.
Formally, given a batch of $N$ referring expressions, BoxRecall@k with IoU threshold $\tau$ is defined as:
\begin{equation}
\resizebox{0.92\linewidth}{!}{$
\text{BoxRecall@}k(\tau) = \frac{1}{N} \sum_{i=1}^{N} \mathbbm{1} \left[ \text{Inst}_i \in \text{Top-}k(i) \cap \text{IoU}(B_i^{\text{pred}}, B_i^{\text{gt}}) > \tau \right]
$}
\end{equation}

In which, $N$ is the total number of queries, $\text{Inst}_i$ denotes the ground-truth object instance for the $i$-th query, and $\text{Top-}k(i)$ is the top-$k$ ranked object candidates sorted by similarity to the query expression. $B_i^{\text{pred}}$ is the predicted bounding box of the matched candidate, and $B_i^{\text{gt}}$ is the ground-truth bounding box of the referred instance. IoU$(\cdot)$ denotes the Intersection over Union between the predicted and ground-truth boxes, and $\tau$ is the threshold controlling the localization strictness. This metric ensures that the model must both retrieve the correct object from a large gallery and accurately localize it. By adjusting the IoU threshold (e.g., 0.5, 0.7, 0.9), we can analyze model performance under varying localization requirements.

\subsection{Implementation Details}
We use the ViT-B and ViT-L variants from SigLIP~\cite{siglip} as our model backbone. For feature extraction, we used a Transformer encoder-decoder architecture following Deformable DETR~\cite{deformabledetr,DN-DETR}, consisting of 6 encoder layers and 6 decoder layers. The number of object queries $M$ is set to 900.
The MORE module uses 1 shared expert for general semantics and a routed expert bank of 4 specialists. Each query activates 2 routed experts via a top-2 gating strategy, enabling targeted relation-aware representation learning.

All experiments are trained on 8 NVIDIA A800 80GB GPUs. We use the AdamW optimizer~\cite{Loshchilov_Hutter_2017} with an initial learning rate of $2 \times 10^{-4}$ and adopt a learning rate schedule that combines linear warm-up with step decay.
During the pretraining stage, the weight decay is set to 0.05, and the per-GPU batch size is 4. In the fine-tuning stage, the weight decay is reduced to 0.0001, and the batch size is increased to 8 per-GPU to improve contrastive learning stability.

\begin{figure}[t]
  \centering
  \includegraphics[width=\linewidth]{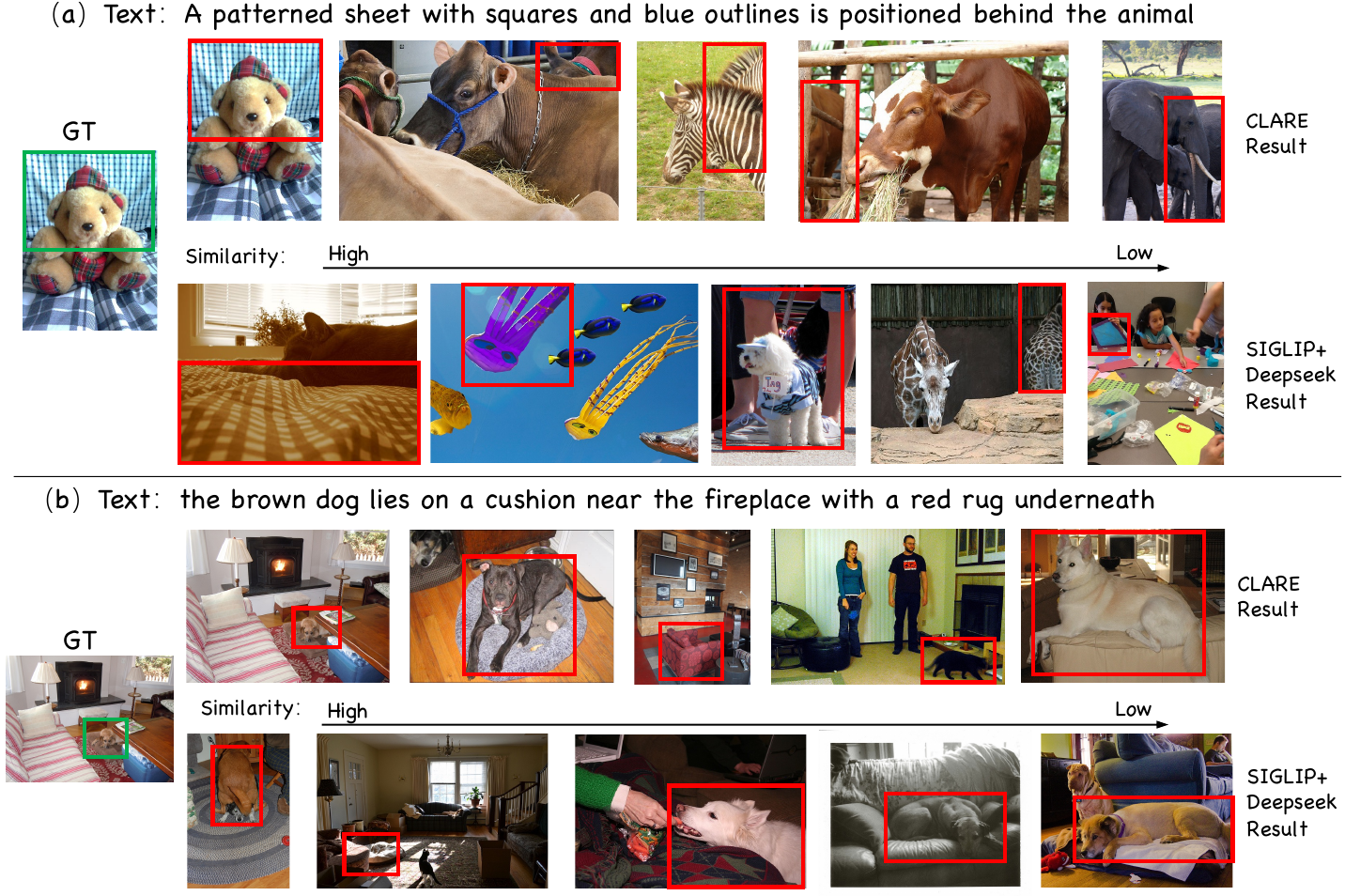}
  \caption{Qualitative comparison between CLARE and a two-stage baseline (SigLIP + DeepSeek-VL2) on REIRCOCO}
  \Description{Comparison of three vision-language datasets.}
  \label{fig:4}
\end{figure}

\begin{table*}[t]
    \centering
    \caption{Performance Comparison of Different Methods on the REC Benchmark}
    \label{tab:2}
    \resizebox{0.8\textwidth}{!}{
        \tiny
        \begin{tabular}{l|c|ccc|ccc|cc}
			\toprule
			\multirow{2}{*}{Models} & \multirow{2}{*}{Backbone} & \multicolumn{3}{c|}{RefCOCO} & \multicolumn{3}{c|}{RefCOCO+} & \multicolumn{2}{c}{RefCOCOg}  \\
			& & val & testA & testB & val & testA & testB  & val-u & test-u \\
			\hline
			\multicolumn{10}{l}{\textbf{Two-Stage}} \\
			\hline
            MAttNet~\cite{mattnet} & RN101 & 76.40 & 80.43 & 69.28 & 64.93 & 70.26 & 56.01 & 66.58 & 67.27 \\
            CM-Att-Erase~\cite{cmatt} & RN101 & 78.35 & 83.14 & 71.32 & 68.09 & 73.65 & 58.03 & 67.99 & 68.67 \\
            DGA~\cite{dga} & VGG16 & - & 78.42 & 65.53 & - & 69.07 & 51.99 & - & 63.28 \\
            RvG-Tree~\cite{Hong_Liu_Mo_He_Zhang_2022} & RN101 & 75.06 & 78.61 & 69.85 & 63.51 & 67.45 & 56.66 & 66.95 & 66.51 \\
            NMTree~\cite{nmtree} & RN101 & 76.41 & 81.21 & 70.09 & 66.46 & 72.02 & 57.52 & 65.87 & 66.44 \\
			\hline
			\multicolumn{10}{l}{\textbf{One-stage}} \\
			\hline
            CRIS~\cite{cris} & RN101 & 70.47 & 73.18 & 66.10 & 62.27 & 68.06 & 53.68 & 59.87 & 60.36  \\
            LAVT~\cite{lavt} & Swin-B & 74.46 & 76.89 & 70.94 & 65.81 & 70.97 & 59.23 & 63.34 & 63.62  \\
            TransVG++~\cite{transvg} & ViT-B & 86.28 & 88.37 & 80.97 & 75.39 & 80.45 & 66.28 & 76.18 & 76.30  \\
            MDETR~\cite{mdetr} & ViT-B & 86.75 & 89.58 & 81.41 & 79.52 & 84.09 & 70.62 & 81.64 & 80.89  \\
            Dyn.MDETR~\cite{Shi_Gao_Huang_Wang_2022} & ViT-B & 85.97 & 88.82 & 80.12 & 74.83 & 81.70 & 63.44 & 72.21 & 74.14  \\
            SimVG~\cite{simvg} & ViT-B & 87.07 & 89.04 & 83.57 & 78.84 & 83.64 & 70.67 & 79.82 & 79.93  \\
			\hline
			\multicolumn{10}{l}{\textbf{MLLM}} \\
			\hline
            shikra-7B~\cite{shikra} & - & 87.01 & 90.60 & 80.20 & \textbf{81.60} & \underline{87.40} & 72.10 & 82.30 & 82.20 \\
            Ferret-7B~\cite{ferret} & - & 87.49 & \underline{91.35} & 82.45 & \underline{80.78} & 87.38 & \underline{73.14} & \underline{83.93} & 84.76 \\
            InternVL2-8B~\cite{chen2024internvl} & - & 87.10 & 91.10 & 80.70 & 79.80 &\textbf{ 87.90} & 71.40 & 82.70 & 82.70 \\
            \hline
            CLARE (ours) & ViT-B & \underline{90.22} & 91.18 & \underline{88.74} & 78.53 & 81.60 & 72.49 & 83.30 & \underline{84.91} \\
            CLARE (ours) & ViT-L & \textbf{91.40} & \textbf{91.95} & \textbf{89.98} & 80.11 & 83.25 & \textbf{74.43} & \textbf{86.30} & \textbf{86.70} \\
			\bottomrule
	    \end{tabular}
    }
\end{table*}

\subsection{Results on REIR}
Since no prior method is designed to directly solve the REIR task, we construct competitive baselines by combining existing Text-Image Retrieval (TIR) and Referring Expression Comprehension (REC) models into a two-stage pipeline. We compare them with our proposed one-stage model, CLARE.

\textbf{Why REC alone is not applicable.}
Although REC models are effective at within-image object localization, they are fundamentally unsuitable for REIR. Directly applying an REC model to the entire gallery requires pairing each query with every image and running $N \times N$ forward passes. For example, on the 5k-query and 5k-image split of REIRCOCO, this results in 25 million inference calls, making large-scale evaluation computationally infeasible. More importantly, REC models lack a cross-image ranking mechanism. They do not output comparable scores across images, making it impossible to sort the gallery or determine which image most likely contains the referred instance. Therefore, REC cannot serve as a standalone baseline for REIR.
\textbf{Two-stage TIR + REC strategy.}
To address the above limitations, we construct a two-stage pipeline that first uses a TIR model to retrieve the most relevant images for each query, and then applies an REC model to localize the referred object within the retrieved candidates. This setup greatly reduces the number of forward passes to $2 \times N$ and is more computationally viable.
\textbf{Baselines.}
We evaluate three strong TIR models: CLIP~\cite{clip}, EVA-CLIP~\cite{evaclip}, and SigLIP~\cite{siglip}. For REC, we use DeepSeek-VL2~\cite{deepseekvl2}, a general-purpose vision-language model, and SimVG~\cite{simvg}, a state-of-the-art REC expert. We evaluate combinations of these TIR and REC models as two-stage baselines.

\textbf{Comparison with CLARE.}
As shown in Table~\ref{tab:1}, our proposed \textbf{CLARE} outperforms all combinations of two-stage baselines across all IoU thresholds and ranking levels. Unlike the cascaded nature of two-stage methods, where failure in retrieval or localization can lead to total failure, CLARE unifies instance retrieval and localization in a single model. This end-to-end design enables more accurate cross-modal alignment and avoids cumulative errors, resulting in both better performance and lower inference cost.

\textbf{Qualitative Analysis.}
Figure~\ref{fig:4} shows qualitative comparisons between our proposed \textbf{CLARE} model and a strong two-stage baseline composed of SigLIP and DeepSeek-VL2. The examples are sampled from the REIRCOCO test set and highlight challenging referring expressions that involve spatial relations, interactions, or multi-object reasoning(e.g., \textit{the brown dog lying on a cushion near the fireplace}).
The baseline often retrieves globally similar but incorrect images due to its lack of instance-level alignment. In contrast, CLARE accurately retrieves and localizes the target object by leveraging cross-image contrastive learning and relation-aware text encoding.
These results qualitatively validate the effectiveness of our end-to-end instance-level alignment strategy and highlight the advantages of modeling fine-grained semantics and inter-object relationships directly during retrieval.

\subsection{Results on REC}
We evaluate CLARE on three standard REC benchmarks: RefCOCO, RefCOCO+, and RefCOCOg. Results are shown in Table~\ref{tab:2}.
Surprisingly, although CLARE adopts a two-stream design, it performs on par with or better than many strong one-stream and MLLM-based models. We attribute this to our cross-image instance-level contrastive alignment, which allows object features to be compared across the entire batch. This encourages each instance to align closely with the referring expression not only semantically, but also in terms of position, size, and contextual relationships.

CLARE achieves particularly strong performance on RefCOCO and RefCOCOg, where relational and spatial expressions are common and well-aligned with our model’s strengths. On RefCOCO+, where such cues are intentionally minimized, our performance is slightly lower. We believe this is because the limited relational information weakens the advantage of our MORE module and contrastive learning design. Nonetheless, CLARE remains competitive even under these conditions, demonstrating its robustness and generalization ability. These results confirm that our method effectively supports both REIR and REC tasks.

\subsection{Results on TIR}

To assess the generalization ability of CLARE beyond REIR, we evaluate it on a standard text-to-image retrieval (TIR) setting adapted to our benchmark. Specifically, we consider a retrieval correct if the retrieved image contains the referred object instance, aligning better with the instance-level nature of our expressions.

As shown in Table~\ref{tab:3}, CLARE achieves substantial improvements over strong TIR baselines such as CLIP, EVA-CLIP, and SigLIP. For example, CLARE-ViT-L achieves 36.40\% R@1, significantly surpassing the best-performing baseline SigLIP (23.25\%). At R@10, CLARE reaches 75.99\%, over 20 points higher than the baselines. These results suggest that CLARE is trained with instance-level supervision and explicitly aligns textual descriptions with object features, making it better suited for real-world applications where users often describe a specific object.

\begin{table}[t]
\centering
\caption{TIR Performance Comparison on the REIRCOCO}
\label{tab:3}
\begin{tabular}{lccc}
\toprule
Model & R@1 & R@5 & R@10 \\
\midrule
OpenAI CLIP         & 15.28 & 31.70 & 40.72 \\
OpenAI CLIP (ViT-L) & 16.79 & 33.80 & 42.43 \\
EVA CLIP            & 18.57 & 36.12 & 45.19 \\
EVA CLIP (ViT-L)    & 20.47 & 38.83 & 47.77 \\
SigLIP              & 21.73 & 40.74 & 49.97 \\
SigLIP (ViT-L-384)  & 23.25 & 43.10 & 52.01 \\
\midrule
CLARE (ViT-B) & \underline{35.68} & \underline{64.36} & \underline{74.49} \\
CLARE (ViT-L) & \textbf{36.40} & \textbf{65.76} & \textbf{75.99} \\
\bottomrule
\end{tabular}
\end{table}

\begin{table}[t]
\centering
\caption{Ablation study of Two-Stage Training. Results are BR@k under IoU threshold $\tau=0.5$.}
\label{tab:4}
\resizebox{0.75\linewidth}{!}{
    \begin{tabular}{ccccc}
    \toprule
    Pretrain & Finetune & BR@1 & BR@5 & BR@10 \\
    \midrule
    \checkmark &          & 4.02  & 12.17 & 18.01 \\
               & \checkmark & 13.44  & 30.95 & 39.83 \\
    \checkmark & \checkmark & 26.39 & 46.44 & 53.28  \\
    \bottomrule
    \end{tabular}
    }
\end{table}

\begin{table}[t]
\centering
\caption{Ablation study on MORE. Results are BR@k under IoU threshold $\tau=0.5$.}
\label{tab:5}
\resizebox{0.75\linewidth}{!}{
    \begin{tabular}{ccccc}
    \toprule
    Total & Selected & BR@1 & BR@5 & BR@10 \\
    \midrule
    0 & 0 & 26.39 & 46.44 & 53.28 \\
    2 & 1 & 26.95 (+0.56) & 46.98 (+0.54) & 53.96 (+0.68) \\
    4 & 1 & 27.65 (+1.26) & 47.88 (+1.44) & 53.80 (+0.52) \\
    4 & 2 & 28.47 (+2.08) & 49.50 (+3.06) & 56.85 (+3.57) \\
    \bottomrule
    \end{tabular}
}
\end{table}

\begin{table}[t]
\centering
\caption{Ablation study on CILA. Results are BR@k under IoU threshold $\tau=0.5$.}
\label{tab:6}
\resizebox{0.75\linewidth}{!}{
\begin{tabular}{lccc}
\toprule
Setting & BR@1 & BR@5 & BR@10 \\
\midrule
without CILA    & 4.74  & 14.07 & 21.24 \\
mini batch CILA     & 18.66 & 39.13 & 48.62 \\
CILA (Full Method) & 26.39 & 46.44 & 53.28 \\
\bottomrule
\end{tabular}
}
\end{table}

\begin{table}[t]
\centering
\caption{Ablation study on CILA loss functions. Results are BR@k under IoU threshold $\tau=0.5$.}
\label{tab:7}
\resizebox{0.75\linewidth}{!}{
\begin{tabular}{lccc}
\toprule
Loss Function & BR@1 & BR@5 & BR@10 \\
\midrule
Focal Loss    & 26.09 & 45.35 & 52.04 \\
SigLIP Loss   & 26.39 & 46.44 & 53.28 \\
\bottomrule
\end{tabular}
}
\end{table}

\subsection{Ablation Study}

We conduct ablation studies using CLARE-ViT-B on the REIRCOCO test set and report BoxRecall@k under an IoU threshold of 0.5. The results verify the effectiveness of our key design components.

\textbf{Effect of Training Strategy.}
To assess the effect of our two-phase training pipeline, we compare four settings: (1) only pretrained, (2) directly finetuned on REIRCOCO without pretraining, and (3) our staged setup, first pretrained on object grounding datasets, then finetuned on REIRCOCO. As shown in table~\ref{tab:4}, both pretraining and REIRCOCO fine-tuning are essential. Pretraining provides grounding capabilities, while REIRCOCO tuning enables instance-level cross-image alignment. The full staged training achieves the best results, confirming the benefit of progressive supervision.

\textbf{Effect of the MORE Module.}
We analyze the impact of different expert routing configurations in MORE. The number of shared experts is fixed to 1, while the number of routed experts and active selection varies. The table~\ref{tab:5} shows that enabling routed experts leads to steady performance gains. Specifically, using 4 routed experts and selecting the top-2 per query gives the best result (56.85 BR@10), confirming that relation-aware routing enhances expression-level adaptation. In contrast, disabling routed experts (i.e., 0 selected) results in 53.28 BR@10. This demonstrates that modeling diverse relational cues, such as spatial positioning, co-occurrence patterns, or relative size, is essential for resolving fine-grained expression ambiguities. See the Appendix for more details.

\textbf{Effect of the CILA size.}
We further examine the design of the CLIA module in table~\ref{tab:6}. As the results show, removing CLIA significantly degrades performance (BR@10 drops to 21.24), indicating the necessity of contrastive supervision. Applying CLIA within mini-batches improves retrieval (BR@10 = 48.62), but still falls short of the full version. Our full CLIA setting achieves the best results (BR@10 = 53.28), confirming that cross-image and cross-batch instance alignment is crucial for learning robust and discriminative visual-textual representations. It enables each object instance to be contrasted against a diverse set of negatives, leading to better generalization in large-scale retrieval scenarios.

\textbf{Effect of Loss Functions in CLIA.}
We further examine different loss functions used within the CLIA module to guide instance-level alignment. As shown in Table~\ref{tab:7}, we compare two alternatives: Focal Loss and SigLIP Loss. Note that this setting only affects the CLIA objective, while the intra-image classification branch remains supervised with standard Focal Loss. We observe that using the SigLIP loss outperforms Focal Loss across all recall levels. These results validate our design choice in extending the SigLIP loss alignment, which is crucial for robust REIR performance.

\section{Conclusion}
We analyze the limitations of traditional TIR and REC tasks in handling instance-level expressions under real-world retrieval scenarios. To address this, we define a new benchmark, Referring Expression Instance Retrieval (REIR), which requires jointly retrieving and localizing object instances from an image gallery. To support research on REIR, we construct a large-scale dataset, REIRCOCO, and propose a baseline model, CLARE, which aligns visual and textual modalities via cross-image instance-level contrastive learning. CLARE achieves state-of-the-art results on REIR and shows strong generalization to existing TIR and REC tasks. REIR expands the scope of vision-language research toward fine-grained, instance-level visual understanding.

\bibliographystyle{ACM-Reference-Format}
\bibliography{sample-base}

\newpage
\clearpage
\appendix

\section{Training}
\subsection{Training Process}
The detailed training settings for CLARE are summarized in Table~\ref{tab:train}. Our training procedure consists of two main stages: (1) grounding pretraining and (2) REIR-specific finetuning. In each stage, we adopt the StepLR scheduler, where the learning rate decays by a factor of 10 at scheduled steps.

\textbf{Stage 1: Grounding Pretraining.}  
To equip the model with basic object grounding capabilities, we first pretrain CLARE on a combination of object detection and referring expression comprehension datasets. Specifically, we use MSCOCO~\cite{COCO} for detection supervision, and RefCOCO~\cite{refcoco}, RefCOCO+~\cite{refcoco}, and RefCOCOg~\cite{RefCOCOg} for REC supervision. We train for 12 epochs in total, with the learning rate dropped after 10 epochs. Multi-scale training is enabled during this stage: images are resized so that the shortest side is between 480 and 800 pixels, while the longest side does not exceed 1333 pixels. AdamW is used as the optimizer with a base learning rate of $5\times10^{-5}$ and weight decay of $0.05$.

\textbf{Stage 2: REIR Finetuning.}  
We then finetune CLARE on the REIRCOCO dataset for 20 epochs. Each training batch samples 8 images with 16 object instances and their corresponding referring expressions. We retain the same multi-scale strategy as in pretraining, and continue using the StepLR scheduler. The model is trained with both retrieval loss ($\mathcal{L}_\mathrm{retrieve}$) and localization loss ($\mathcal{L}_\mathrm{box}$), with loss weights empirically set to 1.0 and 5.0, respectively. For efficient training on large-scale data, we follow Detic~\cite{Detic} to adopt multi-task sampling across GPUs—each GPU processes different subsets of referring expressions per iteration.

\textbf{Additional Techniques.}  
To ensure stability and consistency across image domains, we apply standard data augmentations including random horizontal flipping, color jitter, and scale jittering. Mixed precision training is enabled to reduce the memory footprint. During training, we also construct negative pairs by sampling mismatched expression-instance combinations to enhance contrastive discrimination. The total training takes approximately 100K iterations using 8 NVIDIA A100 GPUs with a batch size of 64.

\subsection{Loss Functions}
\label{loss-func}
For training the proposed CLARE model on the REIR task, we adopt a combination of retrieval and localization losses to jointly optimize cross-modal instance alignment and spatial localization.

\underline{\bm{$\mathcal{L}_\mathrm{focal}$}}.  
To supervise the instance-text matching, we adopt a Focal Loss-based formulation for the retrieval objective. Given the raw similarity score $s$ between a textual query and a visual instance, the matching probability $p$ is computed as $p=\sigma(s)$, where $\sigma$ denotes the sigmoid function. The retrieval loss is then formulated as:
\begin{equation}
\mathcal{L}_\mathrm{focal}(\pt) = - \alpha_t (1 - \pt)^\gamma \log (\pt),
\end{equation}
where 
\begin{equation}
\pt=
\begin{cases}
p &\text{if the instance matches the expression,} \\
1 - p &\text{otherwise}.
\end{cases}
\end{equation}
We set $\gamma = 2$ and $\alpha_t = 0.25$ following~\cite{RetinaNet}.

\underline{\bm{$\mathcal{L}_\mathrm{box}$}}.  
For precise instance localization, we use a combination of GIoU loss and $\ell_1$ loss as in DETR-based frameworks~\cite{DETR,deformabledetr}:
\begin{equation}
\mathcal{L}_\mathrm{box}(b,\hat{b}) = \lambda_{giou}\mathcal{L}_\mathrm{giou}(b,\hat{b})+\lambda_{L_1}\Vert b-\hat{b} \Vert,
\end{equation}
\begin{equation}
\mathcal{L}_\mathrm{giou}(b,\hat{b})=1 - \mathrm{IoU}(b,\hat{b}) + \frac{A^c(b,\hat{b}) - U(b,\hat{b})}{A^c(b,\hat{b})},
\end{equation}
where $A^c(b,\hat{b})$ is the area of the smallest enclosing box, and $U(b,\hat{b})$ is the union area of $b$ and $\hat{b}$.

The final training objective is a weighted sum:
\begin{equation}
\mathcal{L} = \mathcal{L}_\mathrm{retrieve} + \mathcal{L}_\mathrm{box}.
\end{equation}

\begin{table*}[t]
\small
    \centering
\resizebox{1.0\linewidth}{!}{
\begin{tabular}{c|c|ccccc|cccc}

\toprule
Stage & Task & Dataset & Sampling Weight & Batch Size & Short & Long & Num GPU & Lr & Max Iter & Step \\

\midrule
\multirow{1}{*} {pretraining}& \multirow{1}{*} {OD\&REC} & COCO~\cite{COCO} RefCOCO~\cite{refcoco} & 1 &  8 & $480\sim800$ & 1333 & 8 & 0.0002 & 60000 & 48000 \\
\midrule
\multirow{1}{*} {finetune}& \multirow{1}{*} {REIR} & REIRCOCO & 1 &  4 & $480\sim800$ & 1333 & 8 & 0.0002 & 60000 & 48000 \\
\bottomrule
\end{tabular}}
    \caption{
    Ingredients and hyper-parameters for our training.
    } 
    \label{tab:train}
\end{table*}

\section{Datasets}
\label{B}
\subsection{Origin Datasets}

\textbf{RefCOCO/+/g.} RefCOCO~\cite{refcoco,refcoco2} contains 142,210 referring expressions, 50,000 referred objects, and 19,994 images. The testA set primarily describes people, while the testB set mainly describes objects other than people. 

RefCOCO+ contains 141,564 expressions, 49,856 referred objects, and 19,992 images. RefCOCO+ referring expressions focus more on attributes of the referent, such as color, shape, and digits, and avoid using words indicating absolute spatial location. 

RefCOCOg [47; 48] is divided into two partitions: the google split and the umd split. Each split includes 95,010 referring expressions, 49,822 referred objects, and 25,799 images.

\textbf{MS COCO Dataset}
The Microsoft Common Objects in Context (MS COCO) dataset~\cite{mscoco} is a large-scale image dataset developed by Microsoft to advance research in object recognition within complex, real-world scenes. It contains over 330,000 images, with more than 200,000 of them annotated in detail. These annotations span 80 object categories and 91 “stuff” categories, comprising over 1.5 million labeled instances. Each image contains an average of 7.7 object instances, making the dataset particularly suitable for tasks that require understanding of rich contextual information. MS COCO supports multiple computer vision tasks, including object detection, instance segmentation, keypoint detection, and image captioning, and is widely used as a benchmark for evaluating model performance in real-world settings.

\textbf{The MS COCO Captions} dataset is a curated subset of MS COCO specifically designed for image captioning tasks. It provides five human-written captions for each image in the training and validation splits, totaling over 1.5 million captions. In addition, the dataset includes a “c40” subset, where each image is associated with 40 diverse captions, aiming to improve evaluation robustness. An official evaluation server is provided with support for standard captioning metrics such as BLEU, METEOR, ROUGE, and CIDEr, establishing MS COCO Captions as a standard benchmark for image-to-text generation.

\textbf{The MS COCO Detection} subset focuses on object detection, offering precise bounding box annotations and category labels for the 80 object classes. Compared to other datasets, COCO Detection images exhibit more cluttered scenes, higher object density, and a greater proportion of small objects, making the detection task significantly more challenging. Evaluation is conducted using Average Precision (AP) and Average Recall (AR) across multiple Intersection-over-Union (IoU) thresholds, providing a comprehensive assessment of detection accuracy.

\subsection{REIRCOCO Construction}
\begin{figure*}[t]
  \centering
  \includegraphics[width=\linewidth]{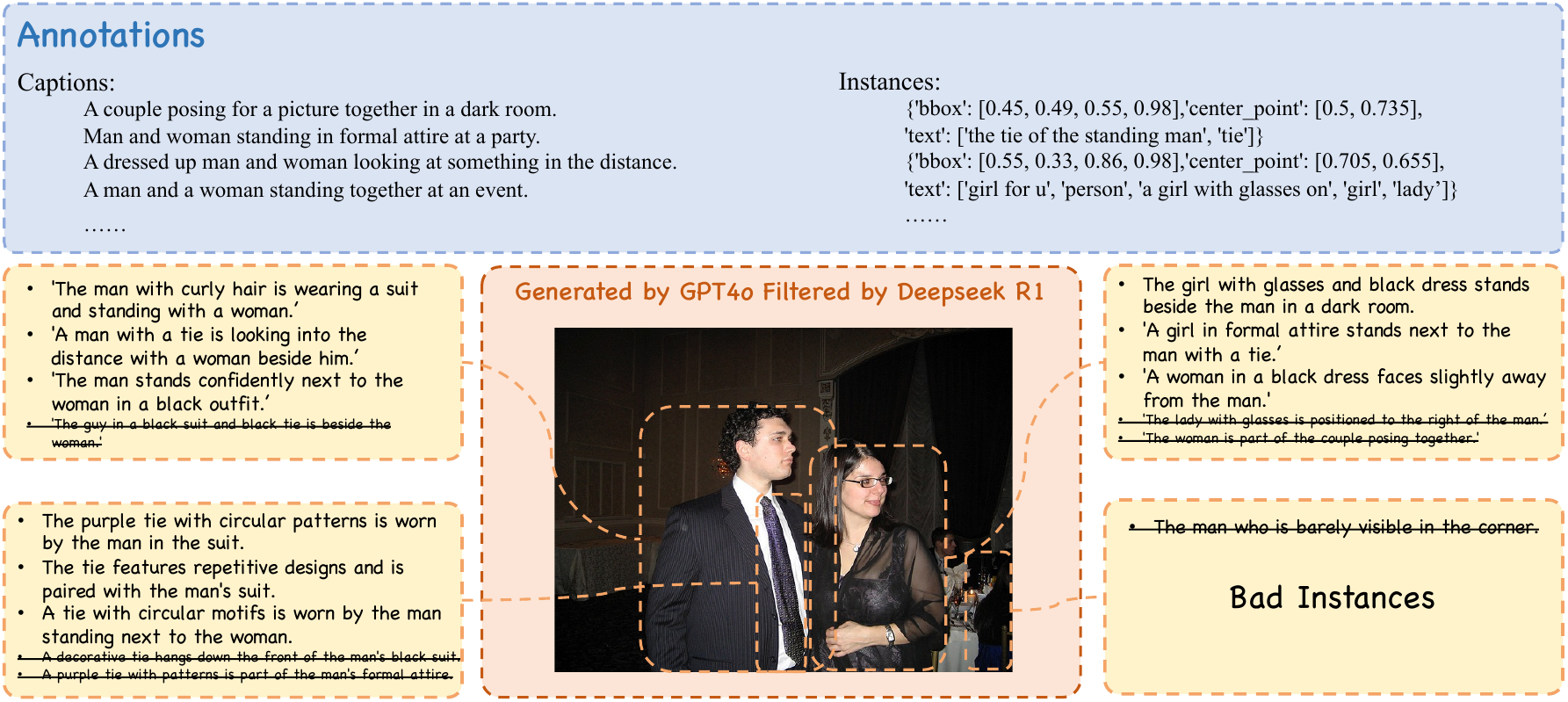}
  \caption{The data construction process for all instances in a single image}
  \Description{Comparison of three vision-language datasets.}
  \label{fig:dataset-create}
\end{figure*}
\textbf{GPT4o. }
To construct high-quality referring expressions for the REIRCOCO dataset, we utilize \texttt{GPT-4o} to generate fine-grained descriptions for each object instance. The generation process is guided by a carefully designed prompt that ensures \textbf{accuracy}, \textbf{uniqueness}, and \textbf{referential clarity}. Given an image and its associated annotation data—including object instances with bounding boxes and image-level captions—the model is instructed to generate \textbf{five distinct natural language descriptions} for each object.

Each generated description:
\begin{itemize}
    \item Relies solely on visual and semantic cues explicitly present in the image and annotations;
    \item Is designed to uniquely identify the target object among many similar instances across different images;
    \item Emphasizes actions, relationships, and spatial configurations, while avoiding direct coordinate references;
    \item Follows a concise, natural, human-like style with short sentences and minimal punctuation;
    \item Encourages diversity without referencing prior descriptions or repeating identical patterns.
\end{itemize}

For instances that are heavily occluded, blurry, or too small to be recognized due to a minimal bounding box, the description \texttt{"Instance quality is poor."} is returned. This generation pipeline enables REIRCOCO to support precise instance-level retrieval and grounding by providing \textbf{rich, contextualized, and discriminative} textual descriptions.

\begin{figure*}[t]
  \centering
  \includegraphics[width=\linewidth]{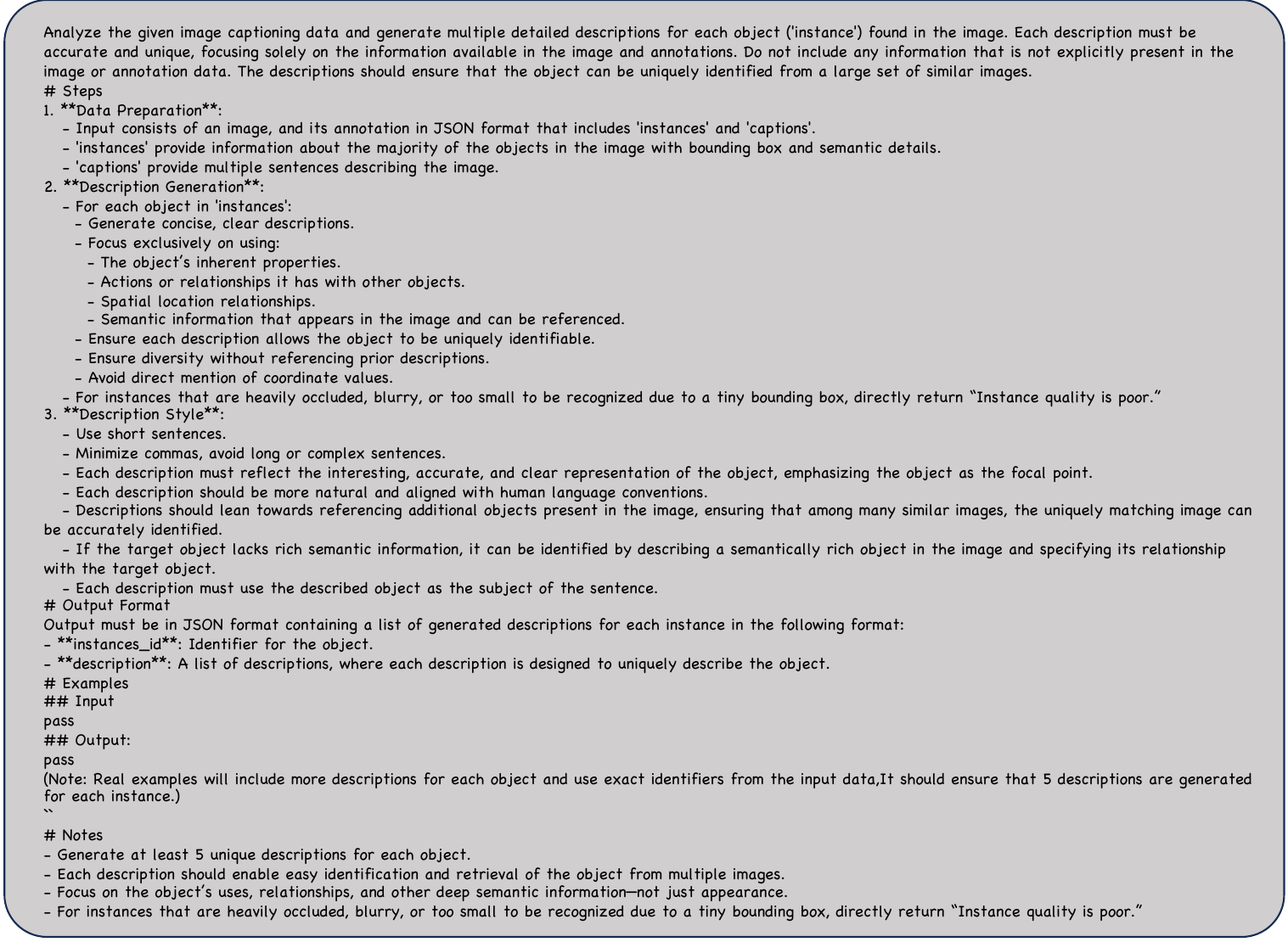}
  \caption{Prompt used for generating textual descriptions with GPT-4o}
  \Description{Comparison of three vision-language datasets.}
  \label{fig:dataset-creation}
\end{figure*}

See Figure~\ref{fig:dataset-creation} for an illustration of the dataset generation process.

\textbf{Deepseek R1.}
After generating candidate descriptions for each object instance using GPT-4o, we adopt a filtering stage based on DeepSeek-VL R1 to ensure the quality, uniqueness, and referential clarity of the descriptions. The filtering process is guided by a structured prompt, which instructs the model to verify whether each generated description is accurate, unambiguous, and grounded solely in the provided image and annotations.

\textbf{Filtering Protocol.}
\begin{itemize}
    \item Each input consists of an image, a set of object instances with bounding boxes and semantic labels (in JSON format), and multiple candidate descriptions for each instance.
    \item The model is prompted to evaluate whether each description:
    \begin{itemize}
        \item Is factually accurate based on the image and annotation.
        \item Uniquely identifies the target object instance among many similar instances.
        \item Avoids hallucinated content or unverifiable assumptions.
        \item Uses the target object as the subject of the sentence.
        \item Follows a natural and concise linguistic style.
    \end{itemize}
    \item Descriptions that are vague, ambiguous, redundant, or based on content not explicitly present in the image are discarded.
    \item If the object is heavily occluded, blurry, or too small to be reliably described, the model returns the placeholder sentence: \texttt{"Instance quality is poor."}
\end{itemize}

\textbf{Output Format.}  
The final output remains in JSON format, with each valid instance entry containing an \texttt{instances\_id} field and a list of verified high-quality \texttt{description}s. Each instance is guaranteed to have up to five unique, human-like descriptions that are referentially discriminative and semantically grounded.

\begin{figure*}[t]
  \centering
  \includegraphics[width=\linewidth]{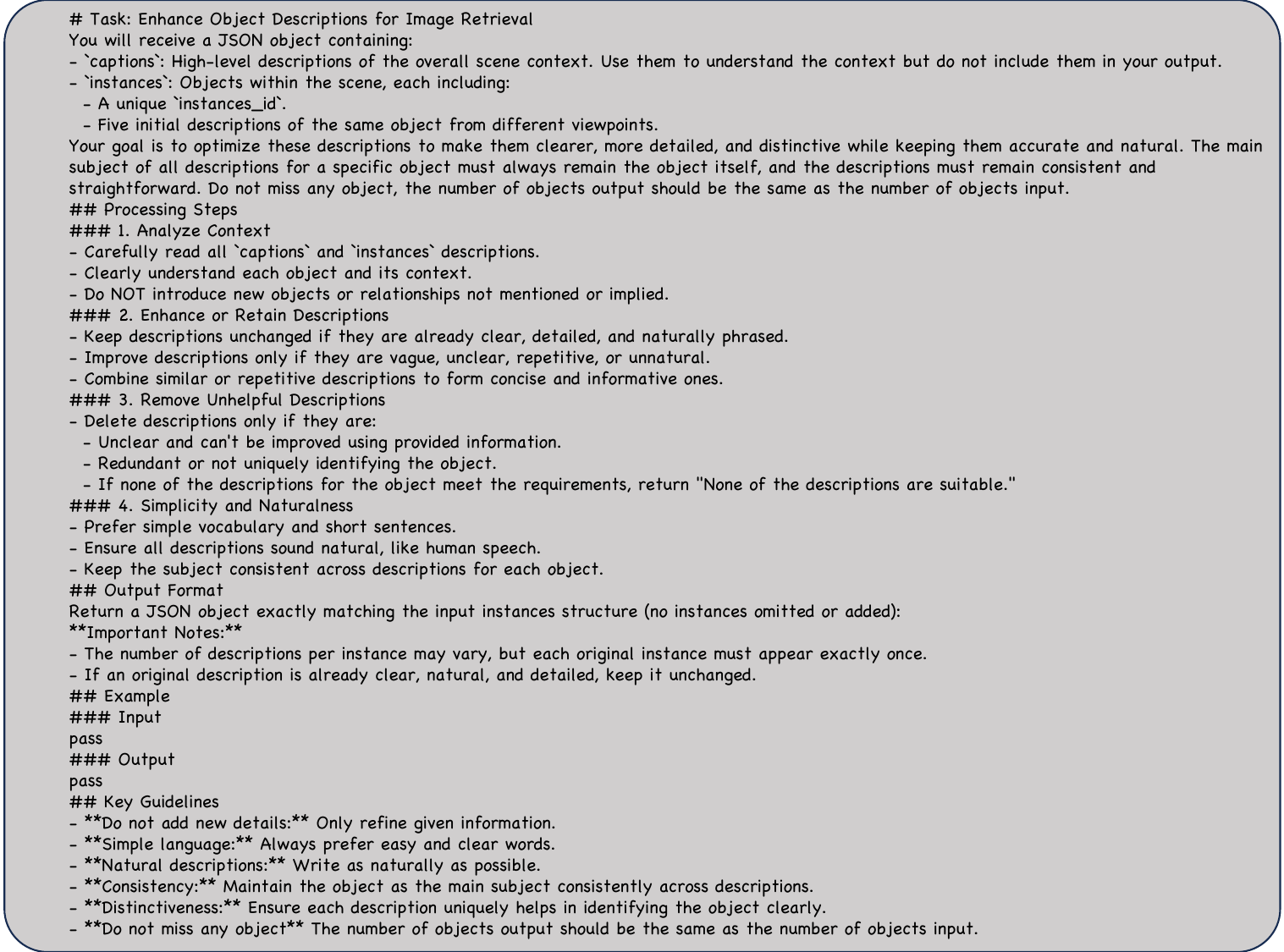}
  \caption{Prompt used for filtering textual descriptions with DeepSeek R1}
  \Description{Comparison of three vision-language datasets.}
  \label{fig:dataset-filter}
\end{figure*}

This filtering stage significantly improves the reliability and precision of the REIRCOCO dataset by eliminating noisy or underspecified referring expressions.
See Figure~\ref{fig:dataset-filter} for an illustration of the dataset generation process. Although DeepSeek cannot accept images as input, it possesses strong reasoning capabilities. Relying solely on captions and GPT-generated descriptions, it can effectively perform filtering.

\subsection{REIRCOCO Visualization}
Due to space limitations in the main text, we present here the full generation process of all instances for a single image, as shown in the figure~\ref{fig:dataset-create}

\section{REIR Visualization}
\subsection{CLARE Result}
\begin{figure*}[t]
  \centering
  \includegraphics[width=\linewidth]{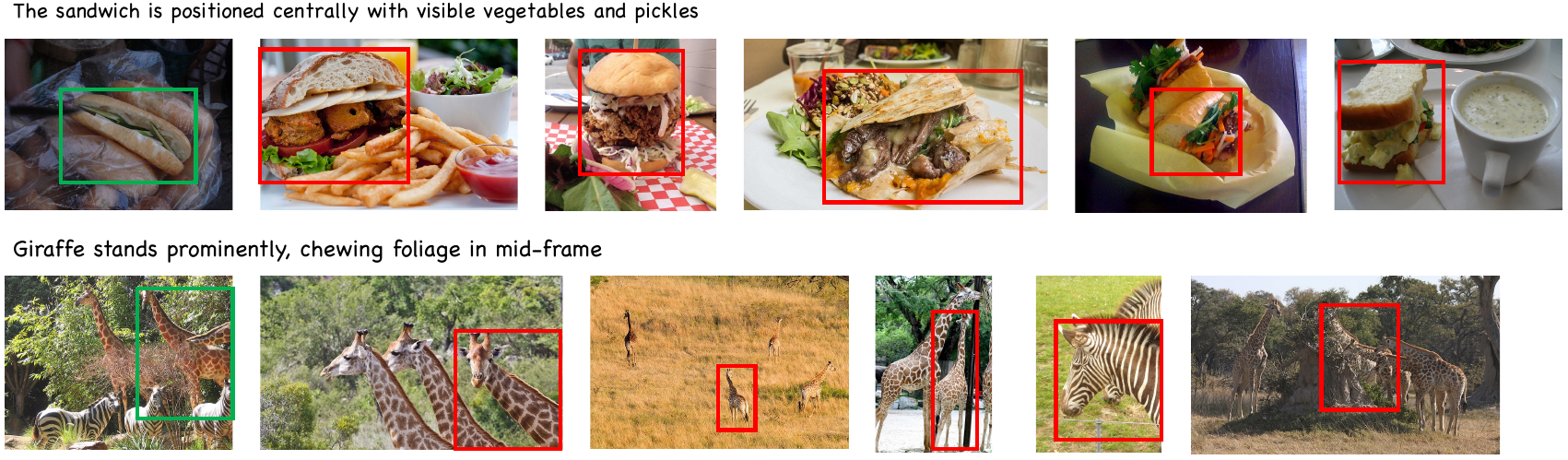}
  \caption{Some failure cases of CLARE on the REIRCOCO dataset}
  \Description{Comparison of three vision-language datasets.}
  \label{fig:bad_case}
\end{figure*}

Qualitative Analysis of Failure Cases.
Figure~\ref{fig:bad_case} presents representative failure cases of our CLARE model on the REIRCOCO test set. Most errors occur when the referring expression involves complex relations, fine-grained distinctions, or ambiguous linguistic constructs. For example, queries such as “the boy sitting behind the girl wearing a red hat” or “the leftmost chair next to the broken table” require the model to jointly understand spatial layouts, relative positions, and object interactions across cluttered scenes.

We observe that CLARE occasionally retrieves a visually similar instance but misinterprets relational cues, leading to incorrect localization. These errors often arise in scenes with multiple similar objects or when relation keywords are subtle or implicit. Additionally, in cases where the expression refers to rare object configurations not well-covered in training (e.g., overlapping small objects), the model may produce high similarity scores for incorrect regions.

These observations highlight the challenge of grounding relational and compositional semantics in open-world settings and suggest future work on enhancing reasoning over multi-object contexts and rare cases.

We also present a number of qualitative examples. Since these examples are composed from images directly during testing, some of them may appear slightly misaligned or irregular.

\begin{figure*}[t]
  \centering
  \includegraphics[width=\linewidth]{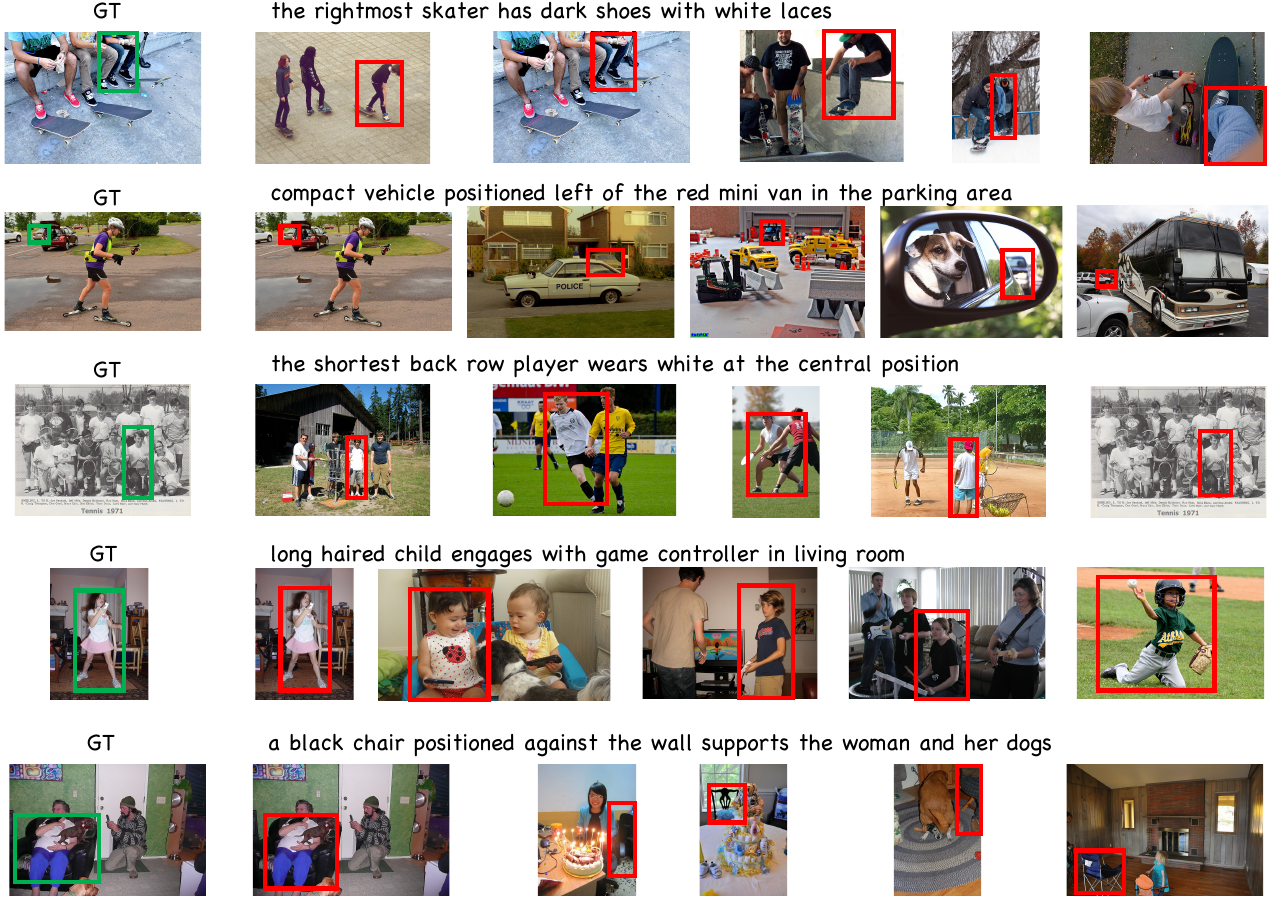}
  \caption{The qualitative results of CLARE on REIRCOCO. This result demonstrates that CLARE can effectively retrieve and localize target objects from image galleries based on textual queries.}
  \Description{}
  \label{fig:more_case}
\end{figure*}


\end{document}